\documentclass{article}
\usepackage[utf8]{inputenc}
\usepackage{authblk}
\usepackage{setspace}
\usepackage[margin=1.25in]{geometry}
\usepackage{graphicx}
\graphicspath{ {./figures/} }
\usepackage{subcaption}
\usepackage{amsmath}
\usepackage{lineno}
\usepackage{multirow}
\usepackage{rotating}
\usepackage{array,booktabs,threeparttable} 
\usepackage{multirow} 
\usepackage{footnote}
\makesavenoteenv{table}

\usepackage[style=nejm, 
citestyle=numeric-comp,
sorting=none]{biblatex}
\title{Multi-Sensor and Multi-temporal High-Throughput Phenotyping for Monitoring and Early Detection of Water-Limiting Stress in Soybean}

\author[1$\dag$]{Sarah E. Jones}
\author[2$\dag$]{Timilehin Ayanlade}
\author[3]{Benjamin Fallen}
\author[2]{Talukder Z. Jubery}
\author[1]{Arti Singh}
\author[2]{Baskar Ganapathysubramanian}
\author[2]{Soumik Sarkar*} 
\author[1]{Asheesh K. Singh*}

\affil[1]{Department of Agronomy, Iowa State University, Ames, IA, United States}
\affil[2]{Department of Mechanical Engineering, Iowa State University, Ames, IA, United States}
\affil[3]{Soybean and Nitrogen Fixation Research Unit, United States Department of Agriculture - Agricultural Research Service, Raleigh, NC, United States}
\affil[*]{Address correspondence to: singhak@iastate.edu or soumiks@iastate.edu}
\affil[$\dag$]{These authors contributed equally to this work.}

\date{}

\begin{document}

\maketitle

\begin{abstract}
Soybean production is susceptible to biotic and abiotic stresses, exacerbated by extreme weather events. Water limiting stress, i.e. drought, emerges as a significant risk for soybean production, underscoring the need for advancements in stress monitoring for crop breeding and production. This project combines multi-modal information to identify the most effective and efficient automated methods to investigate drought response. We investigated a set of diverse soybean accessions using multiple sensors in a time series high-throughput phenotyping manner to: (1) develop a pipeline for rapid classification of soybean drought stress symptoms, and (2) investigate methods for early detection of drought stress. We utilized high-throughput time-series phenotyping using UAVs and sensors in conjunction with machine learning (ML) analytics, which offered a swift and efficient means of phenotyping. The red-edge and green bands were most effective to classify canopy wilting stress. The Red-Edge Chlorophyll Vegetation Index (RECI) successfully differentiated susceptible and tolerant soybean accessions prior to visual symptom development. We report pre-visual detection of soybean wilting using a combination of different vegetation indices. These results can contribute to early stress detection methodologies and rapid classification of drought responses in screening nurseries for breeding and production applications.
\end{abstract}


\section{Introduction}

Biotic and abiotic stresses, exacerbated by weather events, can lead to billions of dollars in U.S. crop insurance payments, economic loss for farmers, and increased consumer prices \cite{Rod2020}. The impact of drought on major grain crops can be severe; therefore, its patterns are traced globally in relation to yield losses. Across different species, yield loss attributed to drought has been investigated. These losses range from 28 to 74\% depending on various factors \cite{jumrani2018impact}. And a recent study of the magnitude, frequency, duration, and timing of droughts has shown that North America is at high risk for reduced soybean (\textit{Glycine max} [L.] Merr.) yield associated with drought \cite{sant2022}.   

In the 1980s, landrace PI416937 (MG 5) from Japan was observed to wilt  more slowly in the field compared to modern cultivars and had a lower yield penalty under water stress \cite{sloane1990field}. Further studies reported that increased root mass, volume, and density, and larger leaf area, higher nodule count, maintained turgor pressure, aluminum tolerance, water conservation strategies, and limited transpiration rate, among other traits, could confer the slow wilting phenotype \cite{hudak1995vegetative,pantalone1996phenotypic,carter1992soybean,valliyodan2017genetic,fletcher2007transpiration,ye2020importance}. Slow wilting lines have shown lower yield reductions under drought stressed conditions \cite{sloane1990field,ye2020importance}. These lines have contributed to breeding progress, as their progenies have higher-yield under drought conditions \cite{carter1992soybean,zhou2020classification}. Canopy wilting has become a proxy measure for drought tolerance in soybean breeding efforts because of the association between slow wilting phenotype with higher seed yield \cite{pathan2014two,ye2020importance}. Furthermore, it has been found that growth stage effects soybean response to drought stress making growth stage an important trait to consider in drought studies \cite{KpoghomouB.K1990Stds}. Ideally, breeders could simultaneously evaluate cultivars and breeding lines for seed yield and growth stage under drought stress and well-watered environments to fully understand drought response in tested lines. However, in situations where drought-prone land is limited, or in early generations when seed quantity is limited for advanced yield trials, the use of proxy traits such as canopy wilting can be utilized for screening and selecting high-performing lines until full-sized yield trials are possible later in the pipeline. 

Several challenges emerge in phenotyping for canopy wilting in breeding and crop production scenarios. Traditional methods involve visually rating wilting severity. Several methods and scales exist for this visual classification in soybean, including the commonly cited scale from 0 - 100 (0 = no wilting, 20 = slight wilting and leaf rolling at the top of the canopy, 40 = severe leaf rolling at the top of the canopy and moderate leaf wilting throughout the canopy and loss of petiole turgidity, 60 = severe wilting throughout the canopy and loss of petiole turgidity, 80 = severe petiole wilting and dead leaves scattered throughout the canopy, and 100 = plant death \cite{king2009differential,kaler2017genome,chamarthi2021identification}, and the equivalent 0 (no wilting) - 5 (plant death) scale \cite{charlson2009polygenic}. Additional methods include a 1-5 scale (1 = no wilting, 2 = few top leaves showed wilting, 3 = half of the leaves showed wilting, 4 = severe wilting ~75\% of the leaves showed wilting, and 5 = severely wilted) and dividing into two categories of slow wilting = average wilting score $ \leq 2.5 $ or fast wilting = average wilting score $ \geq 2.5 $ used in cases rated on the 1-5 scale \cite{ye2020importance,zhou2020classification}. While integral to breeding programs and crop production, several limitations emerge with visual ratings. Visual ratings can be prone to inter- and intra-rater variation, provide no early warning for farmers equipped with irrigation systems, and are time-consuming to collect. Furthermore, breeding programs evaluate thousands of test lines at multiple locations each year \cite{vieira2021numbers}. Canopy wilting is affected by temporally variable environmental conditions, leading to increasingly severe wilting symptoms as the day progresses. Therefore, it is essential that rapid, automated methods for drought screening be implemented in breeding programs to facilitate selection with increased speed and accuracy \cite{singh_chapter21_2021, Singh2021htp}. The immense scale of breeding programs necessitates rapid phenotyping to decrease time and labor costs, and appropriate statistical analysis that can handle complex data.

Sensors on UAVs offer improved speed and spectral resolution beyond human vision with high applicability in breeding programs \cite{herr2023unoccupied, guo2021uas}. A comprehensive array of sensors, including RGB, multispectral, hyperspectral, and thermal, are readily accessible for integration with UAV flights tailored to many agricultural applications \cite{herr2023unoccupied, singh_chapter28_2021}. The combined speed and spectral density of high throughput phenotyping necessitate machine learning (ML) strategies to detect feature hierarchy and recognize patterns in multi-modal data to provide data-driven solutions \cite{singharti2016machine, SinghAsheeshKumar2018DLfP, Singh2021htp}. Common ML objectives in plant stress phenotyping include identification, classification, quantification, and prediction of plant stress \cite{singharti2016machine}. The ICQP paradigm and ML/DL approaches have been applied to solve issues in crop production and breeding including disease identification and severity classification \cite{Ghosal2018disease, Rairdin2022sds}, insect identification \cite{ChiranjeeviShivani2023Dlpr} and can even be applied to below ground traits \cite{CarleyClaytonN.2023Umle}.

Limited research is available that studied UAV-based methods for canopy wilting assessment in the field. Correlation between soybean wilting and canopy temperature was found in 10 soybean genotypes via a thermal infrared camera mounted on a kite and balloon \cite{bai2018aerial}. Another study used multiple sensors, including RGB, infrared thermal, and multispectral cameras mounted on uncrewed aerial vehicles (UAV) to classify 116 soybean genotypes into two classes of slow vs. fast wilting utilizing four imagery-based features inputted into a support vector machine (SVM) algorithms with an accuracy of 0.80 \cite{zhou2020classification}. Similar to SVMs, the Random forest model, an ensemble learning algorithm, also has been widely used \cite{singharti2016machine} and have been successful in plant stress identification and classification \cite{de2023classification}. In peanut (\textit{Arachis hypogaea}), peanut canopy wilting was classified into six wilt classes using 11 features extracted from aerial-based RGB imagery and logistic regression with an accuracy of 0.69 \cite{sarkar2021peanut}. The same data was used to classify wilting into binary categories of turgid vs. wilted canopy with an increased accuracy of 0.88.

In addition to classification of stress in breeding applications, both breeders and farmers have an interest in the early detection of stress. Early stress detection plays a crucial role in breeding programs by enabling breeders to opportunistically screen or gather stress data, especially when faced with unpredictable environmental stresses during less severe years. Early drought stress detection has not been widely studied in soybean, however early detection has been successful in other crops and stresses. Hyperspectral imaging has been used for early detection of charcoal rot in soybean prior to visual symptom development \cite{nagasubramanian2018hyperspectral}. Artificial neural networks have also been used in the early detection of herbicide injury, classifying 240 genotypes into three classes of tolerant, moderate, and susceptible Dicamba injury responses \cite{vieira2022differentiate}. A study of pine tree bark beetles showed that RGB data alone could not detect early infestations, however, red-edge bands played a significant role in early infection detection in a time series data set \cite{yu2022early}. 
The objectives of this study were: (1) develop a pipeline for rapid classification of soybean drought symptoms and (2) investigate methods for early detection of drought stress. 

\section{Materials and Methods}
\subsection{Field Design}

Field experiments were conducted in a rain-fed drought nursery at the Muscatine Island Research Farm in Fruitland, IA (41°21’N, 91°08’W) on Fruitfield coarse sand in 2022. Two-row, 1.52 m plots with 76 cm row spacing and 0.91 m alleys were planted with a seeding density of 66 seeds per square meter on 28 May 2022. 

The experimental design was a randomized complete block design with three replications of 450 MG 0-IV lines. Test lines included a 31-member subset from the SoyNAM panel \cite{song2017genetic,diers2018genetic}, eight maturity and yield checks, and 411 PI lines from a mini core collection subset from the USDA Soybean Germplasm Core Collection representing the diversity of soybean that can be grown in Iowa. PI lines originate from 28 countries across North America, Europe, Asia, and Africa. 

\subsection{Data Collection }

Data was collected across multiple single-day time points 46 days after planting (DAP), 65 DAP, and 81 DAP in 2022 to investigate stress progression. Entire replications were phenotyped in the shortest possible time and the on same day. Visual canopy wilting was recorded on the equivalent scale of 1-6, quantifying the level of wilting severity seen in the plant canopy (1 = no wilting, 2 = slight wilting and rolling in the top of the canopy, 3 = somewhat severe leaf rolling in the top canopy, moderate wilting of leaves throughout the canopy, some loss of petiole turgidity, 4 = severe wilting of leaves throughout the canopy with advanced loss of petiole turgidity 5 = petioles severely wilted and dead leaves throughout much of the canopy, 6 = plant death \cite{charlson2009polygenic,king2009differential}. One individual collected visual wilt scores between 12:00 and 2:00 p.m. to prevent inter-rater variability within a replication. The average soybean growth stage per plot was recorded \cite{fehr1977stages}. 

All sensor and growth stages data were collected between 10:00 a.m. and 2:00 p.m as shown in Table \ref{tab:sensor_spec}.  

\begin{sidewaystable}
    \caption{List of sensors, sensor specifications, and overview of methodology and data points captured in this study.}    
    \centering
    \begin{tabular}{|>{\raggedright}m{4cm}| >{\raggedright}m{3cm}| >{\raggedright}m{2cm}| >{\raggedright}m{1.75cm}| >{\raggedright}m{1.5cm}| >{\raggedright}m{1.5cm}| m{1.1cm}| m{1.1cm}| m{1cm}|}
            \hline
           \textbf{Sensor} & \textbf{Sensor Specification} & \textbf{Wavelength range (nm)} & \textbf{Spectral resolution} & \textbf{Altitude} & \textbf{Overlap} & \textbf{Geno\-types} & \textbf{Repli\-cations} & \textbf{Time points} \\ \hline

Canon T5i digital SLR camera$^1$ & EF-S 18-55 mm f/3.5-5.6 IS II & 400 - 700 & R,G,B & 1.3 m & N/A & \multirow{6}[80]{*}{450} & \multirow{6}[80]{*}{3} & \multirow{6}[80]{*}{3} \\ \cline{1-6}

ASD FieldSpec 4 Hi-Res$^2$ & hyperspectral reflectance & 350 - 2500 & 1 nm & 1 m above canopy & N/A &&& \\ \cline{1-6}

DJI Phantom 4 Advanced UAV$^3$ & 20 MP, 1-inch CMOS with a 24 mm focal length & 400 - 700 & R,G,B & 30 m & 85\% front, 75\% side &&& \\ \cline{1-6}

Zenmuse X5S + DJI Inspire UAV$^3$& 45 mm & 400 - 700 & R,G,B & 45 m & 70\% front, 80\% side &&& \\ \cline{1-6}

Micasense RedEdge-Mx Dual camera system$^4$ + DJI Matrice 600 Pro UAV$^3$ & 10 band multispectral & 444 - 842 & 444(28), 475(32), 531(14), 560(27), 650(16), 668(14), 705(10), 717(12), 740(18), 842(57) & 30 m & 80\% front, 80\% side &&& \\ \cline{1-6}

FLIR Vue Pro R thermal imager$^5$ + DJI Matrice 600 Pro UAV$^3$ & 9mm focal length (640 x 512 pixels) & 7500 - 13500 & 1 band thermal & 20 m & 80\% front, 80\% side &&& \\ \hline

\multicolumn{9}{|l|}{$^1$Canon USA, Inc., Melville, NY}\\

\multicolumn{9}{|l|}{$^2$Malvern Panalytical, Malvern, United Kingdom}\\

\multicolumn{9}{|l|}{$^3$SZ DJI Technology Co., Ltd., Shenzhen, China}\\

\multicolumn{9}{|l|}{$^4$AgEagle Aerial Systems, Inc., Wichita, KS}\\

\multicolumn{9}{|l|}{$^5$Teledyne FLIR LLC, Wilsonville, OR}\\

\hline
            \end{tabular}

    \label{tab:sensor_spec}
\end{sidewaystable}

\subsection{Data Processing }

Data processing included stitching and orthomosaic generation from UAV images, cropping plot boundaries, background removal, and feature generation from canopy pixels. UAV, RGB, multispectral, and thermal flight data were stitched in Pix4D using GPS coordinates of ground control points in the field. From the resulting orthomosaics, plot boundaries were demarcated and snipped via Python code interfacing with ArcGIS \cite{Carroll24IDC}. 

Vegetation pixels were separated from the soil background, and mean reflectance was extracted from each modality via a modality-dependent pipeline. For RGB images, we separated the vegetation in the foreground from the soil in the background using the HSV (Hue, Saturation, Value) color space. RGB images were transformed to the HSV color space, and the pixels with HSV colors between the range of HSV values from (25, 20, 50) to (80, 255, 255) were kept while the other pixels were masked. This range, gotten through trial and error on our data, represented the contours that contain the pixels defined as vegetation. A similar pipeline was employed for the RGB bands in the multispectral data, and the masks created were used to crop out the vegetation from each of the ten multispectral bands. As the thermal modality did not have an associated registered mask, we used the K-means clustering algorithm to segment each plot and adaptively select the vegetation based on the approximate vegetation-to-soil pixel ratio, which was dependent on the time point. The mean reflectance of all soybean plots for multispectral and thermal imagery was extracted for further analysis, similar to those conducted for the RGB data. To characterize various vegetation characteristics, vegetation indices (VIs) were calculated from the multispectral and hyperspectral data using equations provided in Table \ref{tab:vi_used}. These indices are widely used to quantify plant health, greenness, and other physiological conditions. As shown in \ref{tab:vi_used}, we also created another modality consisting of VIs based on only the visible (Red, green, and blue) bands.

\subsection{Data Analysis}
Data from each sensor was analyzed in two phases. Phase one targeted monitoring drought stress severity and phase two targeted early detection of drought stress prior to visual symptom development.

\subsubsection{Monitoring of Drought Stress}
\begin{figure}
    \centering
    \includegraphics{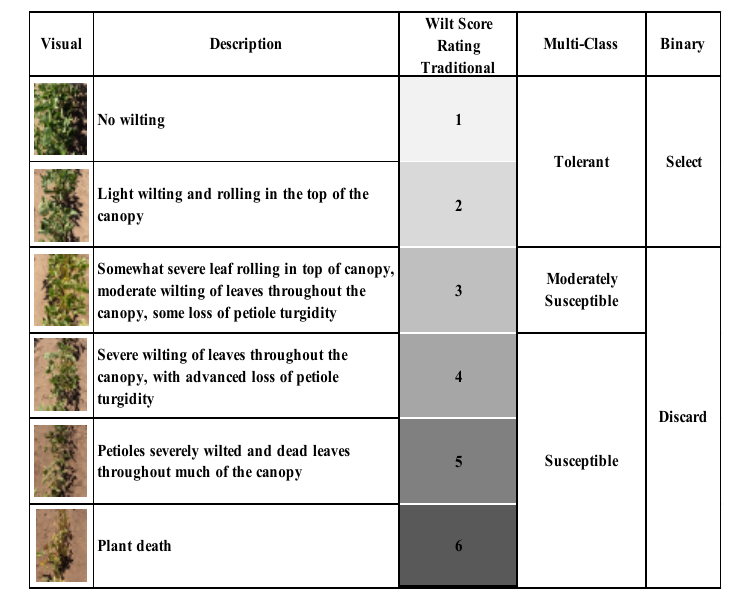}
    \caption{Class categorization of drought stress symptoms in soybean utilized in analysis. Visual ground-truth data was collected on 450 diverse accessions in water-limited screening nursery via traditional wilt score rating scale of 1-6 adapted from previous studies \cite{king2009differential,kaler2017genome,chamarthi2021identification}. Visual plot scores of 1-6 were re-classified into multi-class (3-class) setting and binary (2-class setting) as specified for two analysis pathways.}
    \label{fig:drought_classes}
\end{figure}

For monitoring, we selected time point 2 (65 DAP) for analysis as soybean accessions were in the V9 - R5 growth stage, and all sensor data were available. This time point included 1350 plots, and a balanced subset was chosen by randomly selecting 104 samples per class, which was the number of samples in the least populated class. The traditional wilt score rating on a scale of 1-6 was re-grouped into larger classes that could be applied in a breeding program for advancement decisions. This led to a merger of classes 1 and 2 into tolerant, class 3 as moderately susceptible, and classes 4, 5, and 6 as susceptible, as shown in Figure \ref{fig:drought_classes}. The traditional wilt score rating was also divided into a binary class setting with a merger of classes 1 and 2 into select and a merger of classes 3-6 in discard. 
The analyses explored classification accuracy first with all three classes and again with the two binary classes.  To explore the performance of visual bands from each sensor, the multispectral and hyperspectral sensors have red, green, and blue bands from multispectral and hyperspectral sensors were stacked to create two additional visual-based modalities, as shown in Table \ref{tab:sensor_spec}. 

Random forest (RF) \cite{breiman2001random} models were used to classify the drought susceptibility traits for the various sensors. In this study, we set the number-of-estimator parameter of the RF model to 100 to learn the non-linear classifier. After trial and error, the number-of-estimator was selected to enhance generalization by capturing diverse patterns across each sensor and due to computation. The model performance was evaluated using a 5-fold cross-validation method, where the data is randomly split into 5-folds, and the RF model is trained on 4-folds, while the remaining fold is used for model testing, repeated 5 times. Classification accuracy is calculated using Eq.  \ref{eq:accuracy}. 

\begin{equation}
\text{Accuracy} = \frac{\text{No of samples classified correctly}}{\text{Total No of samples}}
\label{eq:accuracy}
\end{equation}

We used cross-validation as a conservative estimate of the model accuracy \cite{joalland2018aerial}. It is also useful in models dealing with limited sample sizes \cite{james2013introduction}. 

Due to the high dimensionality of the hyperspectral data, with 2151 wavebands. We formulated the problem of identification of the maximally effective waveband combinations among 214 bands as the combination of an optimization and a classification problem using a genetic algorithm as an optimizer and a random forest as a classifier. This subset was created by averaging ten consecutive bands in the hyperspectral data, excluding the first and last bands and including the red, green and blue bands) outlined in \cite{nagasubramanian2018hyperspectral}.

\subsubsection{Early Detection}
For early detection, we use the three time points' data from the 2022 growing season to formulate a time series problem. Time points 1, 2, and 3 represent the V3 - R2, V9 - R5, and R2 - R8 growing stages of the crop, respectively. Time points 1 and 2 are relabeled with the final labels of tolerant or susceptible, as observed in time point 3, so the early time points are paired with their eventual classifications at R2 - R8. We also filtered the reclassified dataset for the extreme classes (tolerant and susceptible) similar to another study \cite{yu2022early}. A balanced set was chosen by randomly selecting 243 samples per class, which was the number of samples in the smallest class. 

Welch's t-test was employed to examine the spectral variations between tolerant and susceptible plots at the various time points to identify the time during the growing season when the classes exhibited distinct reflectance. Additionally, the analysis sought to identify the specific bands contributing to these differences. A RF model was used to evaluate the descriminative ability of the classes at the different time points based on the reclassified dataset above. Multiple RF models were trained using multispectral and hyperspectral data, their VIs,  and combinations of all features.

\begin{sidewaystable}
    \caption{List of vegetation indices (VIs) utilized for drought monitoring and early detection. VIs below the line in the table are based on visible wavelength bands.}
    \centering
    \begin{tabular}{llcc}
        \hline
        Index  & Name & Formula & Reference  \\
        \hline
        NDVI & Normalized Difference Vegetation Index & $\frac{p780 - p670}{p780 + p670}$ & \cite{Rouse1973MonitoringVS} \\
        PRI & Photochemical Reflectance Index & $\frac{p531 - p570}{p531 + p570}$ & \cite{penuelas1995assessment, gamon1997photochemical} \\
        RARSa & Ratio Analysis of Reflectance Spectra A & $\frac{p675}{p700}$ & \cite{chappelle1992ratio} \\
        RARSb & Ratio Analysis of Reflectance Spectra B & $\frac{p675}{p650 \cdot p700}$ & \cite{chappelle1992ratio} \\
        RARSc & Ratio Analysis of Reflectance Spectra Carotenoids & $\frac{p760}{p500}$ & \cite{chappelle1992ratio} \\
        RDVI & Red Normalized Difference Vegetation Index & $\frac{p800 - p670}{\sqrt{p800 + p670}}$ & \cite{roujean1995estimating} \\
        EVI & Enhanced Vegetation Index & $2.5 \cdot \frac{{(p800) - (p670)}}{{(p800) + 6 \cdot (p670) - 7.5 \cdot (pBLUE) + 1}}$ & \cite{huete2002overview}\\
        GCI & Green Chlorophyll Index & $\frac{{(p800)}}{{(p570)}} - 1$ & \cite{esri_band_arithmetic} \\
        MSAVI & Modified Soil-Adjusted Vegetation Index & $\frac{{2 \cdot (p800) + 1 - \sqrt{{(2 \cdot (p800) + 1)^2 - 8 \cdot ((p800) - (p670))}}}}{2}$ & \cite{msavi} \\
        NDRE & Normalized Difference Red Edge & $\frac{{(p790 - p720)}}{{(p790 + p720)}}$ & \cite{SIMS2002337} \\
        RECI & Red-Edge Chlorophyll Index & $\frac{{(p800)}}{{(p740)}} - 1$ & \cite{esri_band_arithmetic} \\
        REV & Red-Edge Vegetation Index & $\frac{{(pp740)}}{{\sqrt{{p670}}}}$ & \cite{red8328017}\\
        ARI & Anthocyanin Reflectance Index & $\frac{1}{{p570}} - \frac{1}{{p740}}$ & \cite{gitelson2009nondestructive}\\
        NDLI & Normalized Lignin Index & $\log\left(\frac{1}{{p1754}}\right) - \log\left(\frac{1}{{p1680}}\right)$ & \cite{serrano2002remote}\\
        NMDI & Normalized Multi-band Drought Index & $\frac{{p860 - (p1640 - p2130)}}{{p860 + (p1640 - p2130)}}$ & \cite{wang2007nmdi}\\
        NWI & Normalized Water Index & $\frac{{p970 - p900}}{{p970 + p900}}$ & \cite{prasad2007genetic}\\
        PSRI & Plant Senescence Reflectance Index & $\frac{{p680 - P500}}{{p750}}$ & \cite{merzlyak1999non}\\
        VREI2 & Vogelmanns Red Edge Index 2 & $\frac{{p734 - p747}}{{p715 + p726}}$ & \cite{vogelmann1993red}\\
        \hline
        RGBVI & Red-Green-Blue Vegetation Index & $\frac{{\text{{p570}}^2 - \text{{p450}} \cdot \text{{p670}}}}{{\text{{p570}}^2 + \text{{p450}} \cdot \text{{p670}}}}$& \cite{mgrvi}\\
        VARI & Visible Atmospherically Resistant Index & $\frac{{\text{{p570}} - \text{{p670}}}}{{\text{{p570}} + \text{{p670}} - \text{{p450}}}}$& \cite{vari} \\
        GLI & Green Leaf Index & $\frac{{2 \cdot \text{{p570}} - \text{{p670}} - \text{{p450}}}}{{- \text{{p670}} - \text{{p450}}}}$& \cite{gli} \\
        MGRVI & Modified Green Red Vegetation Index & $\frac{{\text{{p570}}^2 - \text{{p670}}^2}}{{\text{{p670}}^2 + \text{{p670}}^2}}$& \cite{mgrvi} \\
        ExB & Excess Blue Index & $\frac{{1.4 \cdot \text{{p450}} - \text{{p570}}}}{{\text{{p570}} + \text{{p670}} + \text{{p450}}}}$& \cite{exb} \\
        ExR & Excess Red Index & $\frac{{1.4 \cdot \text{{p670}} - \text{{p570}}}}{{\text{{p570}} + \text{{p670}} + \text{{p450}}}}$ & \cite{exr} \\
        ExG & Excess Green Index & $2 \cdot \text{{p570}} - \text{{p670}} - \text{{p450}}$ & \cite{exg} \\
        ExGR & Excess Green Minus Excess Red Index & $(2 \cdot \text{{p570}} - \text{{p670}} - \text{{p450}}) - \left(\frac{{1.4 \cdot \text{{p670}} - \text{{p570}}}}{{\text{{p570}} + \text{{p670}} + \text{{p450}}}}\right)$ & \cite{exgr} \\
        \hline
    \end{tabular}

    \label{tab:vi_used}
\end{sidewaystable}

\section{Results}

\subsection{Monitoring}

\subsubsection{Classification performance based on Ground Sampling Distance for Visual Features}

\begin{table}[h]
    \centering
    \begin{threeparttable}[t]
\caption{Performance of Random Forest methods to classify drought response in soybean at different ground sampling distances. Results include accuracy obtained from mean reflectance of the visible bands (Mean) and those leveraging Principle Components from PCA \tnote{a}}

\label{tab:gsd}
    \begin{tabular}{cclccc} 
            \hline
           Classes&Data&Sensor&  GSD (cm)& Accuracy &Standard Deviation\\
        \hline
   &&& & &\\
 3& Mean& Hyperspectral& NA& 0.34&0.042\\
 & & Handheld& 0.01& 0.36&0.054\\
 & & Inspire& 0.33& 0.48&0.050\\
 & & Phantom& 0.82& 0.42&0.039\\
 & & Multispectral& 2.0& 0.54&0.040\\
 & & & & &\\
 \hline
 & & & & &\\
 & PCA& Handheld& 0.01& 0.36&0.055\\
 & & Inspire& 0.33& 0.44&0.076\\
 & & Phantom& 0.82& 0.35&0.048\\
 & & Multispectral& 2.0& 0.33&0.053\\
 & & & & &\\
 \hline
 & & & & &\\ 
            
          2&Mean&Hyperspectral&  NA &  0.48&0.025\\ 
            &&Handheld&  0.01& 0.60&0.050\\ 
            &&Inspire&  0.33&  0.68&0.063\\ 
            &&Phantom&  0.82& 0.64&0.041\\ 
            &&Multispectral&  2.0& 0.71&0.086\\ 
   &&& & &\\
         \hline
   &&& & &\\ 
           
           &PCA&Handheld&  0.01&  0.59&0.077\\ 
            &&Inspire&  0.33&  0.63&0.068\\ 
            &&Phantom&  0.82&  0.45&0.063\\
   &&Multispectral& 2.0& 0.53&0.045\\ 
            
   &&& & &\\
         \hline
    \end{tabular}

\begin{tablenotes}
  \item[a] Principal Component Analysis (PCA) is a statistical technique used for dimensionality reduction. Here, PCA with 90\% variance explained was applied to the data.
\end{tablenotes}
\end{threeparttable}
\end{table}

Table \ref{tab:gsd} shows the results of the performance of the RF algorithms for wilt classification for varying ground sampling distances using the mean reflectance of the visible (RGB) bands. The performance of the RF model is reported as the mean accuracy of a 5-fold cross-validation. In addition, we investigated the performance of each ground sampling distance in an increased dimension by leveraging the 90\% Principal Components (PCA) from the flattened 2-dimensional data. For the increased dimension setting, the Inspire sensor (0.33 cm/pixel) outperformed other sensors for classifying wilt stress. Using the mean reflectance for analysis, the Multispectral sensor (2.0 cm/pixel) outperformed other sensors in the multi-class setting. A similar trend can be observed in the binary setting.

\subsubsection{Hyperspectral band selection for Classification}

With the high dimensionality of hyperspectral data and its impact on performance,  it is important to have an efficient analysis pipeline in place for wilt classification. Hence, we determined the minimal number of the most effective hyperspectral wavebands to reduce band correlation, preserve spectral information, and lower the computational costs of working with hyperspectral data. Using the GA-RF algorithm, the performance of the selected combinations of bands was evaluated based on their classification accuracy. A multiclass and binary classification using the selcted bands obtained a classification accuracy of 0.40 and 0.58, with a 5.3\% and 7.1\% increment compared to using all wavebands for the multi-class and binary settings, respectively. Three of the five wavelengths selected were in the near-infrared region (995 - 1940 nm), while other bands were below 995nm.

\subsubsection{Sensor Performance in Wilting Classification}

We analyzed the impact of using different types of sensors on RF models trained on a fixed dataset and evaluated the performance of vegetative indices based on sensors from the multispectral and hyperspectral data. As shown in Table \ref{tab:diffsenseperf}, VIs derived from multispectral sensors had the highest mean accuracy in both multi-class and binary-class settings. Similarly, the multispectral sensor was the second highest-performing sensor, with accuracy close to that of the VIs, thus highlighting the efficacy of the multispectral sensor for classifying wilting in soybeans and the improvement in using VIs compared to using sensors directly. Furthermore, Table \ref{tab:diffsenseperf} shows the improved performance achieved by utilizing the UAV-based multispectral sensors compared to ground-based hyperspectral sensors, which also holds when employing vegetation indices derived from these sensors.

\begin{table}[h]
\caption{Mean Accuracy (Acc.) and Standard Deviation (S.D.) of Different Sensors and Indices. }
    \centering
    \begin{tabular}{lcclll}
     \hline
          Sensor Type&  \multicolumn{2}{c}{3 Classes}&& \multicolumn{2}{c}{2 Classes}\\
 & Mean Acc.& S.D.& & Mean Acc.&S.D.\\
         \hline
          Multispectral Sensor&  0.54& 0.037&& 0.74&0.093\\
          Thermal Sensor&  0.32& 0.060&& 0.47&0.038\\
          Phantom RGB Sensor&  0.42& 0.039&& 0.64&0.041\\
          Handheld RGB Sensor&  0.36& 0.054&& 0.60&0.050\\
          Inspire RGB Sensor&  0.48& 0.050&& 0.68&0.063\\
          Hyperspectral Sensor (GA-RF Selected bands)&  0.40& 0.048&& 0.59&0.048\\
          12 VIs - Multispectral sensor&  0.58& 0.034&& 0.74&0.087\\
          8 rgb-based VIs - Multispectral sensor& 0.47&0.038&&
          0.68&0.039\\
          12 VIs - Hyperspectral sensor&0.38& 0.062&& 0.51&0.089\\
          18 VIs - Hyperspectral sensor&  0.37& 0.058&& 0.56&0.100\\
          \hline
    \end{tabular}
   
    \label{tab:diffsenseperf}
\end{table}

\subsubsection{Performance of Wavebands in Multispectral Modality}

\begin{figure}
  \centering
\includegraphics{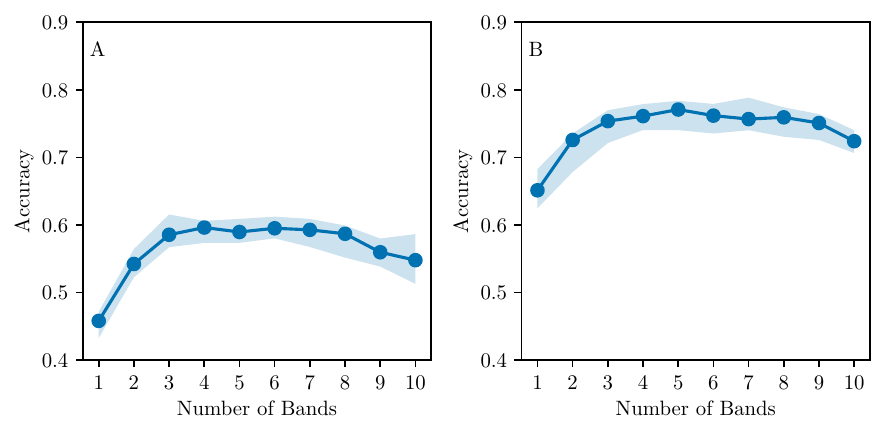}
\caption{Accuracy of random forest method for soybean drought stress using all possible combinations of bands of the multispectral sensor. Mean accuracy for different number of bands for (A) 3 classes and (B) 2 class drought classification.}
\label{fig:mh_combo}
\end{figure}

To investigate the contributing factors for why the multispectral modality outperformed other sensors, we conducted an exhaustive search of all possible combinations of the bands of the multispectral sensor to determine the best combination and number of bands. Figure \ref{fig:mh_combo} a and b show the result of the mean accuracy alongside maximum and minimum performance for each number of bands for the multispectral sensor. The green-531, NIR, red, and red edge band combinations outperformed all combinations with a mean accuracy of 0.60 and 0.77 for the multi-class and binary settings, respectively. In addition, we also investigated the disparity in the various sensing platforms (ground based and UAV imagery). It was observed that ground based RGB imagery had higher mean variances in the red, green, and blue bands (1.01, 1.07, 0.81) compared to the red, green, and blue bands of the multispectral UAV imagery (0.89, 0.75, and 0.72).

\subsubsection{Performance Improvement with multiple sensors}

The combinations of multiple sensor data were evaluated using a backward elimination algorithm. The model using all nine sensor data (phantom RGB, inspire RGB, proximal RGB, Thermal, multispectral, hyperspectral, and vegetative indices from Multispectral and hyperspectral indices, including the RGB-based indices) as predictors (the full model) had an overall classification accuracy of 0.65 in the multi-class setting, the reduced model via backward elimination using a subset of features (phantom, inspire, hyperspectral, thermal, multispectral, and multispectral based vegetative indices) reached the highest overall accuracy of 0.67. For the binary setting, combining all sensors resulted in an accuracy of 0.78. The reduced model, with the Phantom RGB sensor eliminated, reached an improved accuracy of 0.82. 

In addition, we assessed the impact of incorporating soybean growth stage in wilt classification. In the multi-class setting, adding growth stage to the full model does not change the accuracy of the model but maintains accuracy at 0.65. In the binary setting, adding growth stage to the full model increases accuracy from 0.78 to 0.80. Higher accuracy is still achieved by the reduced model that excludes the Phantom RGB imagery. However, including growth stage in the reduced model decreases accuracy from 0.67 to 0.65 in the multi-class setting and from 0.82 to 0.76 in the binary setting. Therefore, the reduced model, excluding the growth stage information, outperformed all configurations.

\subsection{Early detection of Soybean wilt}
\subsubsection{Qualitative Analysis }
In the 2022 growing season, multi-temporal data was collected. Field investigations within the first time point (46 DAP) observed no evident wilting symptoms in susceptible soybean plots. At the second time point (65 DAP), the soybean leaves and petioles began to showcase wilting symptoms in the susceptible plots, which became more evident in the third time point at 81 DAP. Therefore, the UAV-based data we explore contains pre- and post-visual canopy wilting development.

\subsubsection{Spectral-temporal changes}

\begin{figure}
  \centering
\includegraphics{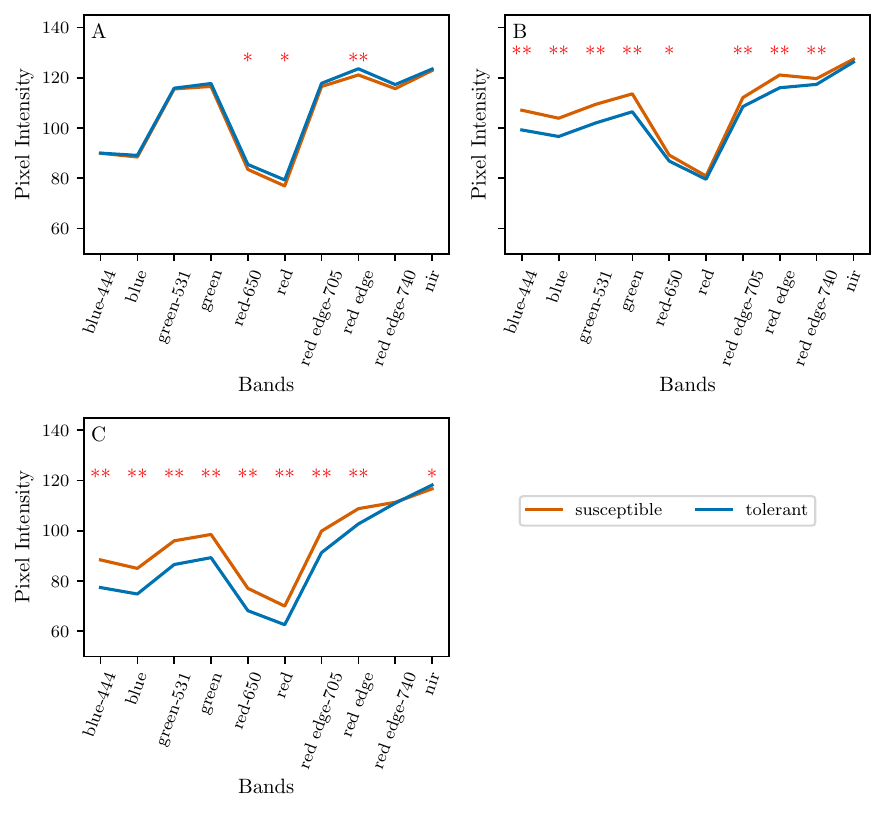}
\caption{Mean spectra of tolerant and susceptible soybean plots of time points (A) Time point 1: 46 DAP, (B) Time point 2: 65 DAP, (C) Time point 3: 81 DAP, at each band of the Micasense RedEdge MX Dual Camera. The symbol * indicates significant differences, with *, **, and *** indicating differences at $p < 0.05$, $p < 0.01$, and $p < 0.001$, respectively.}
\label{fig:three_spectral_graphs}
\end{figure} 

In time point 1, an independent two-sample t-test showed significant differences in the red $(p<0.05)$ to red edge band spectra $(p<0.01)$ of susceptible and tolerant soybean plots, demonstrating the utility of this band measurement in comparison to other bands for classifying soybean plots in the early stages of wilting. For the subsequent time points across the growing stages, the other bands contributed more to the classification (Figure \ref{fig:three_spectral_graphs} B, \ref{fig:three_spectral_graphs} C). Reflectance intensity declined in both groups for the green wavebands and remained steady in the red to infrared wavebands.

\begin{figure}
  \centering
\includegraphics{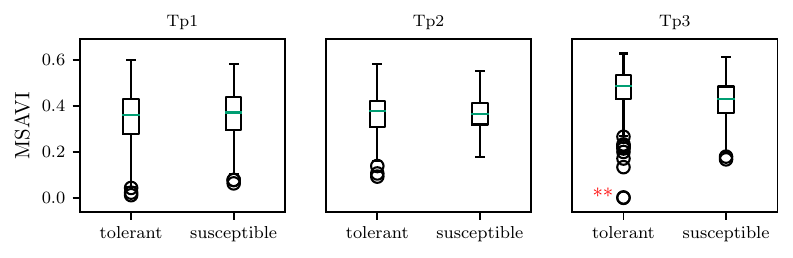}
\includegraphics{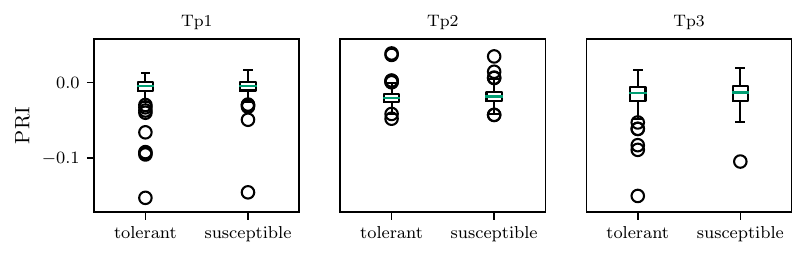}
\includegraphics{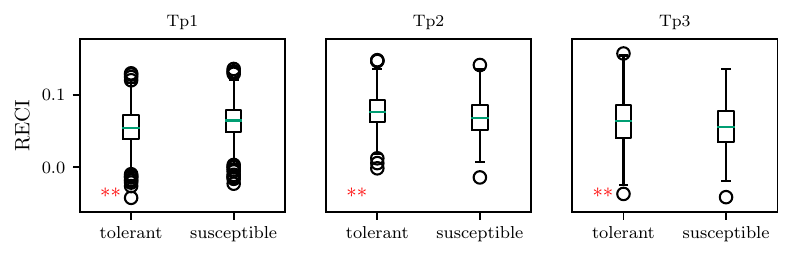}
\includegraphics{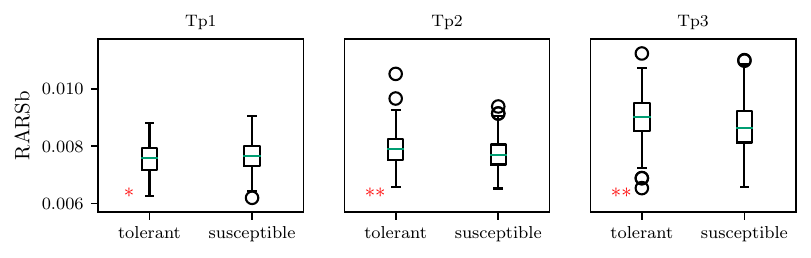}
\caption{Boxplots showing changes in Modified Soil-Adjusted Vegetation Index (MSAVI), Photochemical Reflectance Index (PRI),  Red-Edge Chlorophyll Index (RECI), and Ratio Analysis of Reflectance Spectra B (RARSb) derived from soybean plot spectra across time points 1, 2, and 3. The symbol * indicates significant differences, with * and ** indicating differences at $p < 0.05$ and $p < 0.01$ levels, respectively.}
\label{fig: three_vi_boxplot}
\end{figure}

The sensitive VIs at the early growth stage were  Red-Edge Chlorophyll Index (RECI), and Ratio Analysis of
Reflectance Spectra B (RARSb). Fig \ref{fig: three_vi_boxplot} shows boxplots highlighting the difference between the tolerant and susceptible groups for some of the VIs tested, where RECI remained significant ($p < 0.01$) across all three time points. 

Susceptible and tolerant soybean plots show significant differences in other VIs from time point two.

\subsubsection{Early Detection}
\begin{figure}
  \centering
\includegraphics{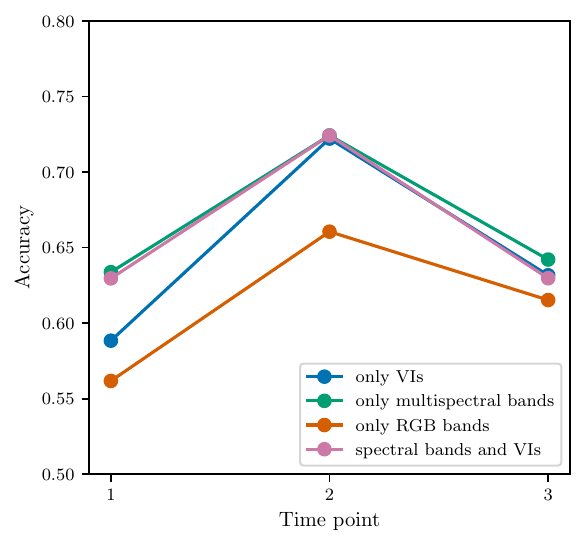}
\caption{Classification accuracy for distinguishing tolerant and susceptible soybean plots by using different types of multispectral-based features over time points 1 (46 DAP), 2 (65 DAP), and 3 (81 DAP).}
\label{fig:rs_fs}
\end{figure}

The RF classification model employed the two best-performing feature types (spectral reflectance and VIs) to separate susceptible and tolerant soybean plots, comparing them to using just the visible bands, which is used by raters. In the first time point, the two groups of soybean plots could not be discriminated efficiently using all features. Nevertheless, the multispectral sensor performed the best with an accuracy of 0.63 (Figure \ref{fig:rs_fs}). 

Over both time points 2 and 3, the wilting classes of soybean plots were separated with an overall accuracy of 0.74. At the last sampling date, the ability to distinguish the two groups diminished to an overall accuracy of 0.64. Among the features, the multispectral sensor performed best in separating the two groups of soybean plots. The classification accuracy was much lower using only mean RGB values compared to other sensors (Figure \ref{fig:rs_fs}).

\section{Discussion}
The efficiency and accuracy of phenotyping protocols are governed by several sensor specifications, including speed of collection, ground sampling distance (GSD), spectral range, and spectral resolution. In examining visual sensor data only (400 - 700 nm), the Inspire UAV with the mid ground sampling distance of 0.33 cm/pixel outperforms the higher ground sampling distance Phantom UAV-based imagery (0.82 cm/pixel) as seen in Table \ref{tab:gsd}. In stress scenarios, lower GSD RGB imagery has been shown to improve the accuracy of stress classification \cite{Carroll24IDC}. The higher resolution may improve accuracy due to smaller canopy sizes under stress and by detecting more data points per leaf, highlighting the importance of high-resolution imagery in stress severity classification.
Interestingly, the Inspire RGB imagery outperformed the handheld RGB imagery (0.01 cm/pixel). The Inspire RGB imaged a larger plot area than the handheld RGB, capturing the full plot compared to a subsection of the plot captured by the handheld RGB. This larger area could have provided a more accurate view of the whole plot's performance and improved accuracy. This is applicable in field plot testing in breeding programs, where small plots have lower accuracy \cite{singh_chapter4_2021}. Furthermore, handheld data collection required 2.5 - 3 hours, while a UAV flight required about 20 minutes. Due to the temporal variation of wilting in response to temperature and solar changes throughout the day, dissimilar collection time duration can introduce higher variance in the ground-based handheld data as observed in this study with higher mean variances in the RGB bands of ground based RGB imagery compared to the RGB bands of the multispectral UAV data, which could affect the accuracy of stress severity classification.

A two-pronged approach to drought phenotyping, including (1) classification of canopy wilting severity and (2) early detection of stress, is highly applicable in both breeding and production environments. In this experiment, three RGB, one multispectral, one thermal, one hyperspectral sensor, and vegetation indices were evaluated for their classification performance in binary and multi-class settings. Overall, in the case of single sensor comparisons, the multispectral and  multispectral based vegetation indices performed the best with a mean binary classification accuracy of 0.74, followed by the Inspire RGB (0.68) between susceptible and tolerant plots. The multispectral bands contributing most to this success included green-531 (14 nm wide centered at 531 nm), red (16 nm wide centered at 650 nm), and NIR (57 nm wide centered at  842 nm). Commonly cited vegetation indices that utilize one or more of these bands explored in this study include NDVI, PRI, RARSa, RARSb, RDVI, EVI, GCI, MSAVI, NDRE, RECI, REV, and ARI  \ref{tab:vi_used}. Higher NDVI values for slow-wilting genotypes have been reported in soybean \cite{zhou2020classification, bai2018aerial} and peanut \cite{balota2017uav}. However, this study explores more sensors and vegetation indices than in previously reported UAV based drought research in soybean. Green-531 was utilized in PRI and shows usefulness in detecting wilting stress due to the morphology of soybean drought response. Improvements in classification accuracy were achieved by employing multiple modalities, which increased classification accuracy in the binary class setting to 0.82.

Ideally, yield, growth stage, and drought stress prediction would be possible simultaneously for application in the breeding program. Prior research has established a correlation between the growth stage of soybeans and their susceptibility to drought stress \cite{KpoghomouB.K1990Stds}. So growth stage data is typically collected and assessed simultaneously with canopy wilting ratings. Despite this, the lack of improvement in the reduced model when including growth stage suggests that growth stage may be inherently captured in the multi-modalities of data. This further indicates that manual growth stage notes may not be imperative for improved classification with this data and model. A few studies in soybean have approached drought phenotyping by predicting yield under drought stress. Nine features from multispectral and RGB-based UAV imagery of 116 soybean genotypes were input into a convolutional neural network to estimate yield under drought stress with an R2 of 0.78 \cite{zhou2021yield}. To test application in a breeding program, a follow-up study showed that 38 features from UAV-based multispectral imagery could be used to classify over 11,000 progeny rows and over 1,000 preliminary yield trial entries into select vs. non-select classes \cite{zhou2022improve}. The model also selected over 60\% of the same selections as a breeder, while increasing average yield of the selected class over breeder selections \cite{zhou2022improve}. This agreement and improved yield show the promise of UAV data and model-based selection methods in the breeding pipeline. Both canopy wilting and yield prediction methods are valuable and offer unique information to plant breeders. Several studies have reported various overlapping and non-overlapping high-performing features in the prediction of soybean yield, including 395 nm, 665 nm, and 675 nm \cite{yoosefzadeh2021application}, NDVI, canopy color, and canopy features \cite{zhou2021yield}, and multiple features including Green, Red Edge, NDVI Red Edge, MSAVI, and NDRE among others \cite{zhou2022improve}. These studies further support the utility of green and red-edge bands. 

Early detection is important for production scenarios where farmers can mitigate drought stress through irrigation and also in breeding programs to identify early symptom development prior to visual wilting in plant canopies. In time point 1, prior to visual symptom development, susceptible and tolerant canopies differed significantly in the red-650, red, and red-edge multispectral bands, showing the importance of multispectral imaging for early detection prior to visual symptom development. Vegetation indices (RARSb, and RECI) were found to promote early detection, with RECI remaining significantly different (p < 0.01) at all three time points pre- and post-visual symptom development. Early drought detection in soybean has not been well studied or reported. However, more work has been done in grapevines where precise water management is crucial for grape and wine quality. One study found that vegetation indices in the red and NIR range, such as NDVI, atmospherically resistant vegetation index (ARVI), enhanced vegetation index (EVI), soil-adjusted vegetation index (SAVI), and optimized soil-adjusted vegetation index (OSAVI), had a higher correlation with leaf water potential compared to other indices, while chlorophyll based indices, especially NDRE, were also highly correlated with leaf water potential \cite{tang2022vine}. In non-drought traits, early detection of stresses in soybean has been reported. For example, in charcoal rot \textit{Macrophomina phaseolina}, a fungal disease promoted by dry environments, early disease detection with an accuracy of 0.97 was made possible by using six selected wavebands, including 475.56, 548.91, 652.14, 516.31, 720.05, and 915.64 nm \cite{nagasubramanian2018hyperspectral}. In a UAV-based study of soybean disease utilizing multispectral data, NDRE was found useful in the early detection of charcoal rot \cite{brodbeck2017using}.

It will be valuable to investigate the role of soil and weather related variables in drought response phenotyping, as it has shown usefulness in yield estimation and prediction \cite{chattopadhyay23yield}.  Drought studies on large genotypic panels can utilize deep learning based methodologies using transfer learning \cite{chiranjeevi2021exploring} and image based phenotyping that has shown finer grained classification of diseases in soybean \cite{Rairdin2022sds}. With finer grained classifications, biotic and abiotic stresses can be combined for meta-analysis \cite{ShookJohnathonM.2021Mfqt} along with traits such as nodulation and rooting depth \cite{ZubrodMelinda2022Ctnt, FalkKevinG.2020SRSA}. This current study utilized high-density spectral and two-dimensional (2D) data to detect and classify drought symptoms. However, to further enhance our understanding, future research could leverage the concept of "canopy fingerprinting" by incorporating three-dimensional (3D) point cloud data. This innovative approach would enable a comprehensive examination of plants response to drought stress in both spectral and 3D changes such as leaf angle and altered volumetric distribution \cite{YoungTherinJ2023ffc, chiranjeevi2021exploring}. By connecting these multi-modal data, researchers can lay the groundwork for developing sophisticated cyber-agricultural systems that seamlessly integrate sensing, modeling, and actuation processes, thereby enhancing our ability to monitor and respond to dynamic environmental conditions in agriculture \cite{Sarkar2023cas}. Early detection methods are particularly of use in cyber-agricultural systems to promote timely actuation.

\section{Conclusion}
Enhancing soybean diversity is imperative, given the current lack of variation in soybean cultivars. Introducing diverse genetic material and wild landraces into breeding populations relies on the ability to make binary selections and eliminations within a breeding program. Leveraging visible and multispectral imagery, particularly in the red-edge and green bands, can facilitate this process. Identifying drought stress at an early stage, even before visible symptoms manifest, relies heavily on the utilization of red-edge bands. This research develops a comprehensive pipeline for integrating data from multiple sensors to classify and detect canopy wilting in soybean. This study emphasizes the significance of high-resolution imaging and the inclusion of both visual and non-visual bands in achieving accurate detection and severity rating.

\section*{Acknowledgments}
The authors would like to acknowledge Brian Scott, Jennifer Hicks and Ryan Dunn for their effort in planting trials in Muscatine. We also thank the many graduate students who assisted in data collection, particularly, Ashlyn Rairdin, Liza Van der Laan, Sam Blair, and Joscif Raigne. We also like to thank Dr. Talukdar Zaki Jubery, Dr. W.T. Schapaugh and Anirudha Powadi for reviewing the paper.

\subsection*{Author Contributions} 
S.J., T.A., S.S., and A.K.S. conceived the project. All authors participated in the project implementation and completion. S.J. planned experiments with A.K.S. S.J. led and completed data collection, imaging, flights and data curation. T.A. developed the machine learning and image analysis pipeline. T.A. and S.J. analysed data and proofed results. A.K.S., S.S., A.S., B.G. contributed to experimentation related resources and overall supervision. S.J. and T.A. wrote the manuscript draft with A.K.S. and S.S. All authors contributed to the final manuscript production.

\subsection*{Funding}
The authors sincerely appreciate the funding support from the United Soybean Board (A.K.S., B.F.), Iowa Soybean Association (A.K.S.), USDA CRIS project IOW04714 (A.K.S., A.S.), AI Institute for Resilient Agriculture (USDA-NIFA \#2021-67021-35329) (B.G., S.S., A.S., A.K.S.), COALESCE: COntext Aware LEarning for Sustainable CybEr-Agricultural Systems (CPS Frontier \#1954556) (S.S., B.G., A.S., A.K.S.), USDA-NIFA FACTS (\#2019-67021-29938) (A.S.), Smart Integrated Farm Network for Rural Agricultural Communities (SIRAC) (NSF S\&CC \#1952045) (A.K.S., S.S.), RF Baker Center for Plant Breeding (A.K.S.), and Plant Sciences Institute (A.K.S., S.S., B.G.). 

\subsection*{Conflicts of Interest}
Authors declare no conflicts of interest with this work.

\subsection*{Data Availability}
Data will be made available upon direct request from Dr. A.K. Singh. singhak@iastate.edu

\printbibliography

\end{document}